\documentclass[review]{elsarticle}

\usepackage{tikz}
\usepackage{graphicx}
\usepackage{amsmath}
\usepackage{hyperref}
\usepackage{booktabs}
\usepackage{caption}
\usepackage{subcaption}
\usepackage{float}
\usepackage{multirow}
\usepackage{tabularx} 
\usepackage[table]{xcolor}  
\usepackage{booktabs}       
\usepackage{geometry}
\usepackage{threeparttable}
\usepackage{array}          

\journal{Computer Methods and Programs in Biomedicine}

\begin{document}

\begin{frontmatter}


\title{The use of vocal biomarkers in the detection of Parkinson's disease: a robust statistical performance comparison of classic machine learning models
}

\author[inst1,inst2]{Katia Pires Nascimento do Sacramento}
\author[inst2,inst3]{Elliot Q. C. Garcia}
\author[inst2]{Nicéias Silva Vilela}
\author[inst4]{Vinicius P. Sacramento}
\author[inst2]{Tiago A. E. Ferreira}

\address[inst1]{Universidade Regional do Cariri, Juazeiro do Norte-CE, Brazil}
\address[inst2]{Universidade Federal Rural de Pernambuco, Recife-PE, Brazil}
\address[inst3]{Futuo-Tech, Recife-PE, Brazil}
\address[inst4]{Universidade Federal do Cariri, Juazeiro do Norte-CE, Brazil}

\begin{abstract}
\textbf{Background and objective:} Parkinson’s disease (PD) is a progressive neurodegenerative disorder that, in addition to directly impairing functional mobility, is frequently associated with vocal impairments such as hypophonia and dysarthria, which typically manifest in the early stages. The use of vocal biomarkers to support the early diagnosis of PD presents a non-invasive, low-cost, and accessible alternative in clinical settings. Thus, the objective of this cross-sectional study was to consistently evaluate the effectiveness of a Deep Neural Network (DNN) in distinguishing individuals with Parkinson’s disease from healthy controls, in comparison with traditional Machine Learning (ML) methods, using vocal biomarkers. \\
\textbf{Methods:} Two publicly available voice datasets were used. Mel-frequency cepstral coefficients (MFCCs) were extracted from the samples, and model robustness was assessed using a validation strategy with 1,000 independent random executions. Performance was evaluated using classification statistics. Since normality assumptions were not satisfied, non-parametric tests (Kruskal–Wallis and Bonferroni post-hoc tests) were applied to verify whether the tested classification models were similar or different in the classification of PD. \\
\textbf{Results:} With an average accuracy of $98.65\%$ and $92.11\%$ on the Italian Voice dataset and Parkinson's Telemonitoring dataset, respectively, the DNN demonstrated superior performance and efficiency compared to traditional ML models, while also achieving competitive results when benchmarked against relevant studies. \\
\textbf{Conclusion:} Overall, this study confirms the efficiency of DNNs and emphasizes their potential to provide greater accuracy and reliability for the early detection of neurodegenerative diseases using voice-based biomarkers.
\end{abstract}

\begin{keyword}
Parkinson’s disease\sep Deep neural networks\sep Machine learning; Vocal biomarkers\sep Early diagnosis\sep MFCC.
\end{keyword}

\end{frontmatter}

\section{Introduction}
Parkinson's Disease (PD) is a progressive neurodegenerative disorder characterized by motor symptoms such as tremors, rigidity, and bradykinesia. In addition, vocal impairments such as hypophonia and dysarthria are prevalent, often occurring in the early stages of the disease. The ability to detect PD using voice signals offers a non-invasive, low-cost, and accessible approach, which is fundamental for aiding in early diagnosis, particularly in low-resource clinical settings.

The use of vocal biomarkers for disease detection through innovative methodology has been consolidated in recent years \cite{Mazur2025, Lefkovitz2024, Tsanas2024, Vasconcellos2023}. The extraction of information, whether by analyzing the behavior of sound propagation or through the intonation and linguistics of the voice, enables the creation of various models capable of detecting alterations that signal potential clinical changes, even before they manifest in the individual. These alterations, detected by biological measures, have several clinical applications, such as for PD, mild cognitive impairment, and Alzheimer's disease, for which machine learning (ML) and deep neural network (DNN) approaches are applied \cite{Chen2022, VasquezCorrea2021, Suhas2024}.

The human voice possesses fundamental characteristics and properties that are influenced by all physiological systems. When one of these systems undergoes a pathological change, the final sound produced varies \cite{Titze2008}. In this light, PD, being a neurodegenerative condition, also influences changes in voice. Therefore, studies using vocal biomarkers have been employed for the early diagnosis of the illness. Innovative strategies utilizing artificial intelligence have been developed to identify PD in its initial stages, including through integrative approaches with multiple parameters \cite{IntegrativeApproachPD}. 

The combination of genetic data and portable devices is already showing notable results, as evidenced in the study \cite{qi2025bidirectional}, which suggests an early detection system through body sensors and genomic information. Alterations in the acoustic characteristics of the voice, such as variations in frequency and intensity, reveal specific acoustic patterns according to the underlying clinical pathology \cite{Li2025}.
The use of machine learning (ML) models for the early identification of Parkinson's disease from speech data has been explored, achieving high predictive accuracy with the Random Forest classification model \cite{GovinduPalwe2023}.

Since these vocal alterations are detectable in the early stages of PD, voice-based diagnostic models have been developed, utilizing biomarkers that allow for the safe differentiation of individuals with PD from those without the disease \cite{LiLeeZhang2019}. The use of artificial intelligence to automatically assess clinical indicators could enable the tracking of disease progression, reduce costs, and increase access to clinical examinations.

Techniques using vocal biomarkers have been improved, thanks to mobile technology solutions. The work of \cite{Zhan2018} proposed a smartphone-based digital scoring system that assesses the severity of Parkinson's disease and allows for monitoring patients at home. The project led by \cite{DeChoudhury2021} collected 10,000 vocalizations of telephone quality and used supervised classification algorithms to detect the severity of PD, as well as cross-validation methods to address overfitting.

In this research, to increase diagnostic accuracy and produce statistically consistent results, a technique was employed that generates 1,000 independent random runs, thus minimizing the variability resulting from splitting the data into training and testing sets and producing performance statistics for each run. The means and standard deviations of these indicators are generated at the end of the process. This approach offers a more rigorous and stable assessment of model performance, and, to date, the authors are unaware of its use in articles related to this project.

Unlike previous studies that employed methods such as k-fold cross-validation \cite{diCesare2024,Hu2020} and bootstrap \cite{neto2024harnessing, karabayir2020gradient, sayed2023arxiv}, this article adopted a strategy very similar to \cite{xu2001monte}, which is random subsampling, known in the literature as an alternative to cross-validation.

Therefore, the objective of this work was to understand the distributions of the performance metrics and, consequently, evaluate the effectiveness of a DNN and traditional ML approaches (like Ramdom Florest, Ligistic Regression, SVM, and Gradiente Boosting) in distinguishing between individuals with PD and healthy individuals, using vocal biomarkers.

The article is organized as follows. Section \ref{sec:MaterialsMethods} presents the datasets and the procedure employed. Section \ref{sec:Results} describes the results obtained and Section~\ref{sec:Discutions} presents the analyses performed based on the results. Finaly, Section \ref{sec:Conclusions}  presents the conclusion of the analysis.

\section{Materials and Methods}\label{sec:MaterialsMethods}

Predictive classification models play a significant role in the early detection of PD, given that its diagnosis is predominantly clinical, involving the evaluation of medical history, symptoms, and neurological examination \cite{johns_hopkins_parkinson_diagnosis}.
In this context, this section provides a detailed description of the methodologies employed in this comparative study, encompassing dataset characteristics, preprocessing techniques, machine learning model specifications, experimental protocols, and statistical analysis procedures.

\subsection{Datasets} \label{sec:data}

Two publicly available voice datasets relevant to the diagnosis of PD were used for a comprehensive evaluation of the model's performance:

\textbf{Voice Dataset of Italian Parkinson's Disease:} this dataset, initially developed by \cite{dimauro2016voxtester} and available at \cite{dimauro2019italian}, also contains voice markers of Italian individuals divided into three main groups: 15 Healthy Young Individuals, 22 Healthy Older Individuals, and 28 Individuals with Parkinson’s Disease. Previous studies in the area, such as \cite{pandey2025parkinson}, \cite{vizza2025through} and \cite{pandey2025cascade}, used this same database.

\textbf{Parkinson's Telemonitoring Dataset:} this dataset was analyzed in \cite{little2007parkinsons}, obtained from the UCI Machine Learning Repository \cite{ref:telemonitoring_dataset}, and comprises biomedical voice measurements of 31 individuals with and without PD, ranging in age from 46 to 85 years. This dataset is employed in many works found in the literature, including \cite{sun2025hybrid}, \cite{saha2025lightweight},  \cite{aliero2025comparative}, among others. 

Both datasets were essential for evaluating the generalizability and robustness of the machine learning models implemented across different data characteristics.

\subsection{Data Preprocessing}

A systematic preprocessing pipeline was applied to the data for each independent experimental run, ensuring consistency and minimizing bias.

\subsubsection{Acoustic Feature Extraction}

The audio samples were standardized with a sampling rate of $16$kHz and adjusted to a $1$-second duration. Subsequently, $13$ \textit{Mel-Frequency Cepstral Coefficients} (MFCCs) were extracted per sample, using the \texttt{torchaudio} library. The parameters included $40$ Mel bands, a $400$ points \textit{Fast Fourier Transform} (FFT), and a step of $160$ samples between the FFT windows. Finally, the MFCCs were calculated for each $25$ms window, and the temporal mean was computed for each coefficient, generating a $13$-dimensional vector for each recording.

\subsubsection{Data Splitting}

The entire MFCC dataset was randomly split for each of the $1,000$ independent experimental runs. The data was divided into two hierarchical phases: first, $80\%$ was allocated to training/validation and $20\%$ to testing. Then, the subset ``training + validation'' ($80\%$ of the total data) was divided into $75\%$ to the training set and $25\%$ to the validation set. 
The sampling method employed was Stratification. It was done uniformly across all sections, thus preserving the data distribution within each subgroup. After that, the data were normalized using z-score standardization (\texttt{StandardScaler}) from the \texttt{scikit-learn} library, in which the means and standard deviations for each attribute of the input vector were adjusted. This z-score procedure was applied independently to the training, validation, and test sets.

\subsubsection{Class Balancing}

During the training process, \textit{oversampling} was applied to minimize the imbalance between the classes (healthy individuals versus Parkinson's patients). This procedure was applied separately to the training and validation sets, avoiding any type of data leakage between the sets. The \texttt{RandomOverSampler} function in the \texttt{balanced-learn} library  was employed here. This technique performs random duplication of samples from the minority class. The goal is to achieve proportionality in the number of samples between the classes in the training set. Consequently, machine learning algorithms tend to exhibit greater generalization capacity and improved performance in identifying relevant patterns.

\subsection{Machine Learning Models}

Preprocessed data were used to train and evaluate the performance of five machine learning models: Random Forest (RF), Logistic Regression (LR), SVM, and Gradient Boosting (GB), commonly used in PD detection. All traditional ML models were implemented in Python via \texttt{scikit-learn} (version 1.5).
The hyperparameter values of the models were tested and also compared with previous studies on PD detection using vocal biomarkers. These values can be found in Table ~\ref{tab:hyperparameters}.

\begin{table}[h!]
\centering
\caption{Main hyperparameter configurations of the machine learning models used in this study.}
\rowcolors{2}{gray!10}{white}
\small 
\begin{tabularx}{0.95\textwidth}{@{}p{3.5cm} X@{}}
\toprule
\textbf{Model} & \textbf{Key Hyperparameters} \\ 
\midrule

\textbf{Random Forest (RF)} & 
\textit{n\_estimators}=100; 
\textit{criterion}=`gini'; 
\textit{max\_depth}=None; 
\textit{max\_features}=`sqrt'. 
 \\[2pt]

\textbf{Logistic Regression (LR)} & 
\textit{solver}=`lbfgs'; 
\textit{max\_iter}=1000; 
\textit{C}=1.0; 
\textit{penalty}=`l2'. \\[2pt]

\textbf{Support Vector Machine (SVM)} & 
\textit{kernel}=`rbf'; 
\textit{C}=1.0; 
\textit{gamma}=`scale'; 
\textit{probability}=True. \\[2pt]

\textbf{Gradient Boosting (GB)} & 
\textit{n\_estimators}=100; 
\textit{learning\_rate}=0.1; 
\textit{max\_depth}=3; 
\textit{subsample}=1.0; 
\textit{criterion}=`friedman\_mse'. \\

\bottomrule
\end{tabularx}
\label{tab:hyperparameters}
\end{table}

Additionally, a Deep Neural Network (DNN) was used to categorize audio recordings, separating healthy individuals from those diagnosed with PD. The DNN architecture was developed following guidelines appropriate for binary classification problems and implemented using the \texttt{PyTorch} library.
The models followed the same data splitting and normalization protocol, being trained and tested in $1,000$ independent runs.

Initial experiments were employed to define the best DNN architecture. The best DNN architecture consisted feedforward neural network fully connected with an input layer composed of 13 neurons and two hidden layers with 64 and 32 neurons, respectively. The ReLU (Rectified Linear Unit) function was applied to both hidden layers. Then, the Dropout regularization technique with a $30\%$ dropout rate was applied after each hidden layer to avoid overfitting \cite{srivastava2014dropout}. Finally, the output layer is composed of one neuron, and its output is processed by a sigmoid function.

The Adam optimizer  was used with a learning rate of $0.003$ and a regularization term $L2$ of $0.001$. The network was trained for $100$ epochs, with a batch size of $32$. Early stopping was applied with a patience of 15 epochs.

\subsection{Statistical Analysis}

Commonly used classification criteria in PD detection analyses --- like  accuracy, precision, recall, F1-Score --- are also used here to evaluate the performance of all models. Data interpretations were performed after $1,000$ independent runs for each model.

The Shapiro-Wilk  and Levene  tests were applied to assess normality and homogeneity of variance in both data sets previously described at the beginning of Section \ref{sec:data}. After analysis, it was found that parametric ANOVA analysis could not be applied to both data sets, since all cited statistical tests' $p$-values were statistically significant $(p=0.0000)$, implying that the null hypotheses (normality and variance homogeneity) were refuted.

Subsequently, the nonparametric Kruskal-Wallis \cite{kruskal1952use} test was used to investigate whether there was a significant difference between the models. The Bonferroni test \cite{bonferroni1936teoria} then analyzed whether there was a significant difference in accuracy between each pair of models.

Box plots were generated from $1,000$ runs of each model on a Pandas DataFrame via \texttt{plotly.express}, helping to visualize model performance.

\subsection{Experimental Protocol}

All models were trained and tested using the same procedure to ensure a fair comparison. Each experiment consisted of $1,000$ independent runs, using the preprocessing, feature extraction, and data splitting described in the previous sections. Random subsampling was applied to generate different training, validation, and test sets for each run.

During training, oversampling was employed to balance the classes in both the training and validation sets separately, thereby preventing data leakage between the sets.

The metrics accuracy, precision, recall, and F1-score were used to measure the performance of each model. For statistical analyses, the Shapiro-Wilk and Levene tests were used to assess normality and homogeneity of variances. Since the normality conditions were not met, nonparametric Kruskal-Wallis tests followed by Bonferroni post-hoc comparisons were applied to determine whether there were differences between the models.

\begin{figure}[htbp]
\centering
\resizebox{0.5\textwidth}{!}{
\begin{tikzpicture}[
    node distance=1.2cm,
    arrow/.style={->, thick},
    base/.style={
        align=center,
        font=\footnotesize,
        rounded corners,
        draw=gray!70,
        minimum width=3.6cm,
        minimum height=0.9cm
    },
    data/.style={base, fill=blue!15},
    process/.style={base, fill=lime!20},
    analysis/.style={base, fill=orange!20},
    result/.style={base, fill=violet!20}
]

\node[data] (start) {Start of Experiment \\ (for each one of the 1000 Independent Runs)};
\node[data, below of=start] (data) {Random Subsampling \\ (Train / Validation / Test)};
\node[process, below of=data] (preprocess) {Preprocessing \& Oversampling \\ (Class Balance)};
\node[process, below of=preprocess] (training) {Model Training \\ (DNN, RF, SVM, LR, GB)};
\node[process, below of=training] (evaluation) {Performance Evaluation \\ (Accuracy, Precision, Recall, F1)};
\node[analysis, below of=evaluation] (stats) {Statistical Analysis \\ (Shapiro–Wilk, Levene, Kruskal–Wallis, Bonferroni)};
\node[result, below of=stats] (end) {Comparison of Models \\ (Consistency \& Reproducibility)};

\draw[arrow] (start) -- (data);
\draw[arrow] (data) -- (preprocess);
\draw[arrow] (preprocess) -- (training);
\draw[arrow] (training) -- (evaluation);
\draw[arrow] (evaluation) -- (stats);
\draw[arrow] (stats) -- (end);

\end{tikzpicture}
}
\caption{Workflow of the experimental protocol used for predictive classification of Parkinson’s disease using vocal biomarkers.}
\label{fig:protocol}
\end{figure}
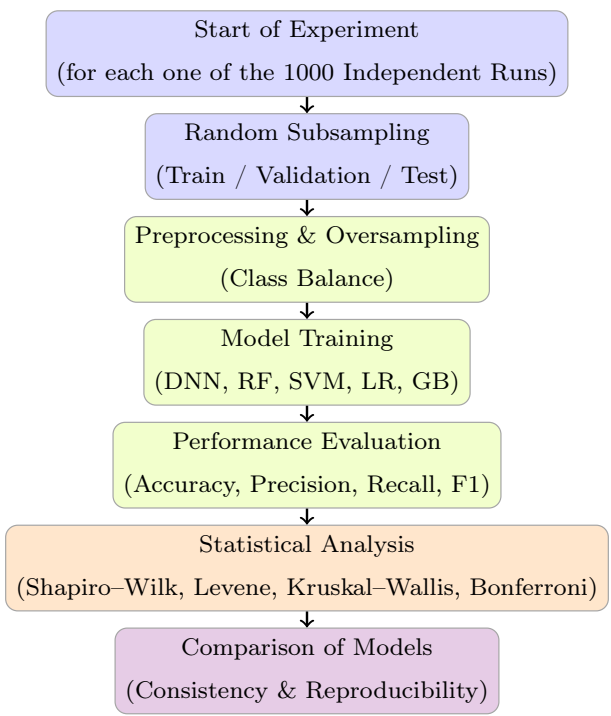

All experiments were performed under the same conditions to maintain consistency and allow for robust comparison of model performance. The flowchart of the experimental protocol is shown in Figure~\ref{fig:protocol}.

\section{Results}\label{sec:Results}

This section presents a detailed analysis, covering classification performance metrics and the results of specific statistical tests, including the Kruskal-Wallis and Dunn (Bonferroni) post-hoc tests. For this purpose, the two publicly available voice datasets relevant to PD detection introduced in Section~\ref{sec:data} were used (Italian Parkinson’s Voice dataset and Parkinson's Telemonitoring UCI).

A key aspect of this study is the rigorous assessment of the analyzed models' robustness and reliability, achieved through 1,000 independent experimental runs for each model. This systematic approach, designed to mitigate random variations in data splits and model initializations, aims to increase the accuracy of performance estimates and the generalizability of derived insights.

The evaluated models comprise a Deep Neural Network (DNN), Random Forest (RF), Logistic Regression (LR), Support Vector Machine (SVM), and Gradient Boosting (GB). Subsequent subsections will detail the distributions of classification accuracies, as shown in Figures~\ref {fig:boxplotital} and \ref{fig:boxplotuci}.

More comprehensive classification metrics, such as precision, recall, and F1-score, are displayed in Table ~\ref {tab:metrics_combined_datasets}. Furthermore, in this table, models sharing the same letter perform statistically equally, based on the results of Dunn's test and the Bonferroni adjustment, which controls for false positives. Therefore, before comparing model performance, rigorous statistical tests were performed to ensure the robustness of the analyses. To this end, normality and homogeneity tests were performed on the precision data, as shown in Table ~\ref{tab:global_tests} and Figure ~\ref{fig:bonferroni}. Additionally, the table details the results of the nonparametric tests mentioned above.

\subsection{Accuracy Distributions}

\begin{figure}[H]
\centering
\includegraphics[width=1.0\textwidth]{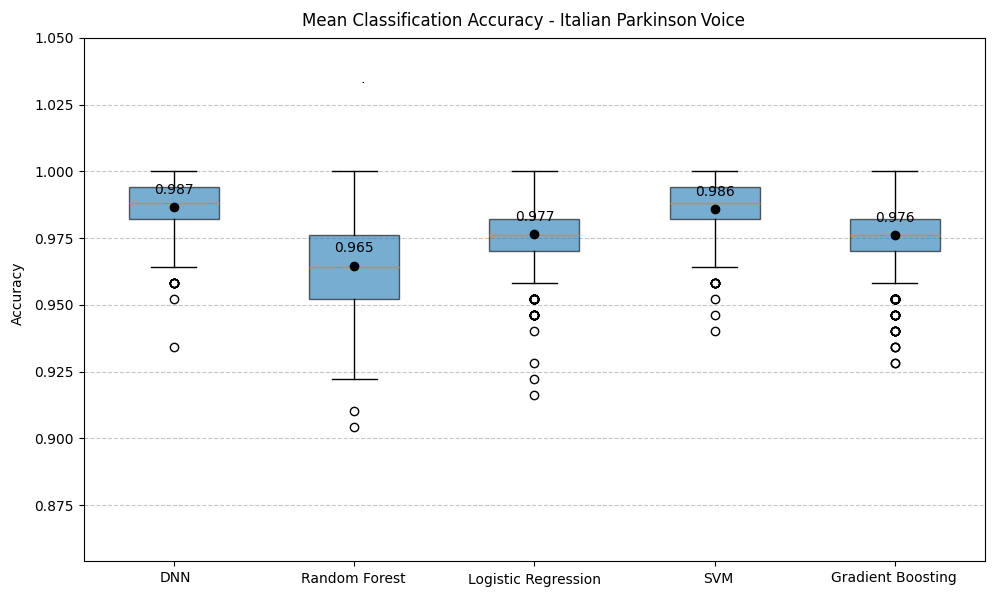}
\caption{Distribution of model performance across 1,000 randomized training runs on the Italian voice dataset.}
\label{fig:boxplotital}
\end{figure}

\begin{figure}[H]
\centering
\includegraphics[width=0.9\textwidth]{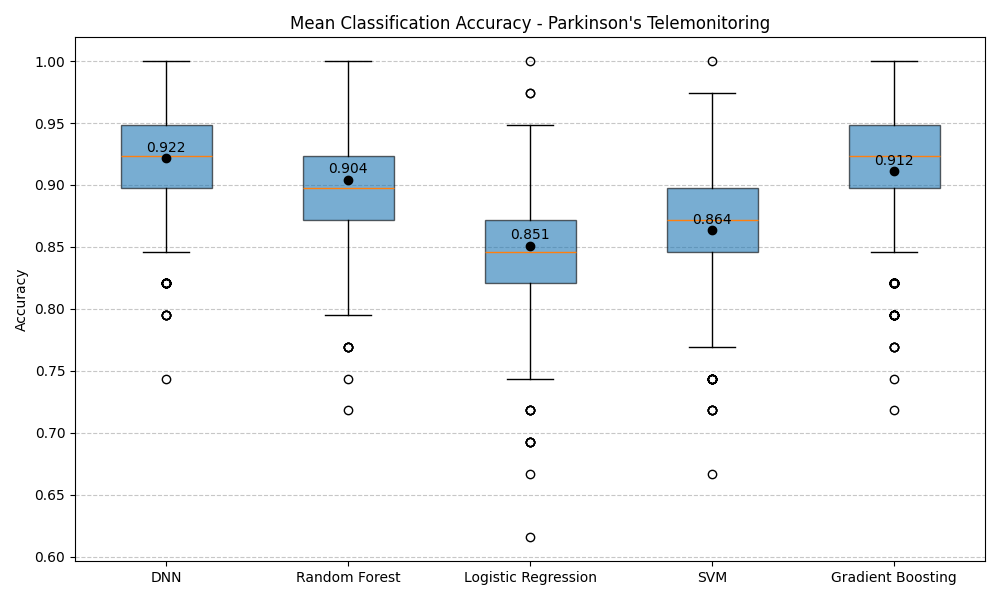}
\caption{Distribution of model performance across 1,000 randomized training runs on the Parkinson’s Telemonitoring dataset.}
\label{fig:boxplotuci}
\end{figure}

Figures ~\ref{fig:boxplotital} and ~\ref{fig:boxplotuci} illustrate boxplots of the mean classification accuracies of five machine learning models evaluated on two publicly available datasets: the Italian Parkinson's Voice and the Parkinson's Telemonitoring UCI. For the Italian dataset, the Deep Neural Network (DNN) and the SVM achieved the highest mean accuracies $(0.9865\pm 0.0091)$ and $(0.9859\pm 0.0093)$, respectively, both with narrow interquartile ranges and minimal outliers. Nonparametric statistical tests (Figure~\ref{fig:bonferroni}) revealed no significant difference between these two models, suggesting that both perform equally well in detecting PD.

On the other hand, the LR $(0.9766\pm 0.0115)$ and GB $(0.9760\pm 0.0124)$ models present wider distributions and lower median accuracy. Their predictive performance is similar, as shown in Figure~\ref{fig:bonferroni}. Despite their high performance, the performance difference between them and the DNN and SVM models is significant.

Although the RF model is a widely used method in biomedical problems, in this dataset, it had the lowest average accuracy $(0.9645\pm 0.0148)$, which is statistically different from the other models, as shown in Figure~\ref{fig:bonferroni}. This difference may be related to the high variability, lower median, and subtle acuities in the data.

According to the performance analysis of average accuracy, in the Parkinson's Telemonitoring dataset, the DNN model also led, achieving a value of $(0.9211\pm0.0150)$. The GB model achieved a very close value of $0.9119$, with a difference of only $0.4\%$. According to the Dunn (Bonferroni) post-hoc test (Figure~\ref{fig:bonferroni}), all distinct pairs of models differed significantly ($p < 0.001$). 

The LR model performed worst, with an average accuracy of $(0.8511\pm 0.0198)$, suggesting less predictive ability compared to the other models.
Overall, for these data, the DNN and GB models showed, on average, superior performance in classifying healthy and diseased individuals.

\subsection{Statistical Significance}

Table~\ref{tab:metrics_combined_datasets} shows the mean values and standard deviations for accuracy, precision, recall, and F1-score across both datasets, allowing for the detection of consistent performance patterns and, consequently, the evaluation of each model's predictive ability.

In the specific context of the Italian Voice Parkinson's disease dataset, as shown in Table~\ref{tab:metrics_combined_datasets}, the DNN demonstrated superiority across almost all metrics: accuracy $0.9865$, precision $0.9850$, recall $0.9895$, and F1-score $0.9872$, all with low variability.

The SVM model achieved a recall of $0.9897$, slightly higher than the DNN. The fact that both classifiers demonstrated greater sensitivity on the Italian Voice data is a relevant finding. In a clinical context, using either method tends to reduce the likelihood of false-negative diagnoses.

However, a model with high sensitivity incurs a precision cost, thus being prone to generating more false positives. Although the DNN and SVM models still have high recall values, they also maintain the best precision values.

LR performed with a precision of $0.9766$, demonstrating a predictive power very different from that observed in the Parkinson's Telemonitoring dataset.

GB and RF achieved precision of $0.9760$ and $0.9645$, respectively. All classifiers in general performed within a range of similar values, demonstrating strong discriminative ability for vocal biomarkers in the Italian voice dataset.

Due to its small standard deviations and, consequently, little variability between runs, the DNN reaffirmed consistency and robustness across all metrics in the Parkinson's Telemonitoring dataset. It presented the highest accuracy $0.9211$, precision $0.9458$, and F1-score $0.9470$ among all the models tested, according to Table~\ref{tab:metrics_combined_datasets}. This high accuracy indicates that when the model predicted an individual with PD, it was correct most of the time. The higher F1-score reinforces that the DNN model achieved the best overall performance.

GB was the second best classifier, with an accuracy of $0.9119$. A recall of $0.9619$ demonstrates its sensitivity in detecting the classes of interest. The RF algorithm achieved a recall of $0.9666$, indicating its high ability to identify PD cases. In a clinical setting, using a classifier with high recall is essential, as a false negative in the case of a serious disease, for example, is very serious. However, the model's high sensitivity may result in lower accuracy $0.9115$ when compared to the DNN.

The RF and GB classifiers showed similar average accuracies, above $0.90$, as shown in Table ~\ref{tab:metrics_combined_datasets}. This similarity suggests that both classifiers are suitable for predicting PD. The SVM, with an average accuracy of $0.903$.

Finally, with the lowest precision $0.8511$, recall $0.9048$, and F1-score $0.9030$, the LR model was the least effective among the models evaluated, suggesting that its simplicity is not suitable for this type of dataset.

The consistency of the DNN's best results across both datasets makes it viable and reliable for complex medical classification problems. While other models performed well, they did not hold up in both contexts.
\begin{table}[H]
\caption{Summary of Classification Metrics (Mean ± Std. Deviation) for Parkinson's Disease Telemonitoring and Italian Voice Datasets. The best metric values are in blodface. For each measure, the underline values indicate that models are statistically equal.}
\label{tab:metrics_combined_datasets}
\setlength{\tabcolsep}{4pt}
\centering
\footnotesize
\begin{tabular}{lcccc}
\toprule
\textbf{Model} & \textbf{Accuracy} & \textbf{Precision} & \textbf{Recall} & \textbf{F1-Score} \\
\midrule
\multicolumn{5}{l}{\textbf{Italian Voice Dataset}} \\
\addlinespace[0.5ex]
DNN\textsuperscript{a} & \textbf{0.9865} ± 0.0091 & \textbf{0.9850} ± 0.0095 & 0.9895 ± 0.0089 & \textbf{0.9872} ± 0.0082 \\
Random Forest\textsuperscript{b} & 0.9645 ± 0.0148 & 0.9506 ± 0.0162 & 0.9837 ± 0.0125 & 0.9667 ± 0.0129 \\
Logistic Regression\textsuperscript{c} & 0.9766 ± 0.0115 & 0.9763 ± 0.0121 & 0.9793 ± 0.0108 & 0.9777 ± 0.0102 \\
SVM\textsuperscript{a} & 0.9859 ± 0.0093 & 0.9835 ± 0.0098 & \textbf{0.9897} ± 0.0087 & 0.9865 ± 0.0085 \\
Gradient Boosting\textsuperscript{c} & 0.9760 ± 0.0124 & 0.9659 ± 0.0137 & 0.9893 ± 0.0101 & 0.9773 ± 0.0109 \\
\addlinespace[1em]
\midrule
\multicolumn{5}{l}{\textbf{Parkinson's Telemonitoring Dataset}} \\
\addlinespace[0.5ex]
DNN & \textbf{0.9211} ± 0.0150 & \textbf{0.9458} ± 0.0198 & 0.9499 ± 0.0185 & \textbf{0.9470} ± 0.0150 \\
Random Forest & 0.9038 ± 0.0162 & 0.9115 ± 0.0223 & \textbf{0.9666} ± 0.0169 & 0.9375 ± 0.0162 \\
Logistic Regression & 0.8511 ± 0.0198 & 0.8783 ± 0.0257 & 0.9311 ± 0.0231 & 0.9027 ± 0.0198 \\
SVM & 0.8638 ± 0.0185 & 0.8756 ± 0.0241 & 0.9551 ± 0.0209 & 0.9124 ± 0.0185 \\
Gradient Boosting & 0.9119 ± 0.0170 & 0.9245 ± 0.0210 & 0.9619 ± 0.0190 & 0.9421 ± 0.0170 \\
\bottomrule
\end{tabular}
\smallskip
\begin{minipage}{\textwidth}
\footnotesize
\textsuperscript{a,b,c} Models with the same letter do not show statistically significant differences $(p > 0.05)$, according to Dunn's post-hoc test. In the Parkinson's Telemonitoring dataset, all models differed statistically.
\end{minipage}
\end{table}

\begin{figure}[H]
\centering
\includegraphics[width=1.0\textwidth]{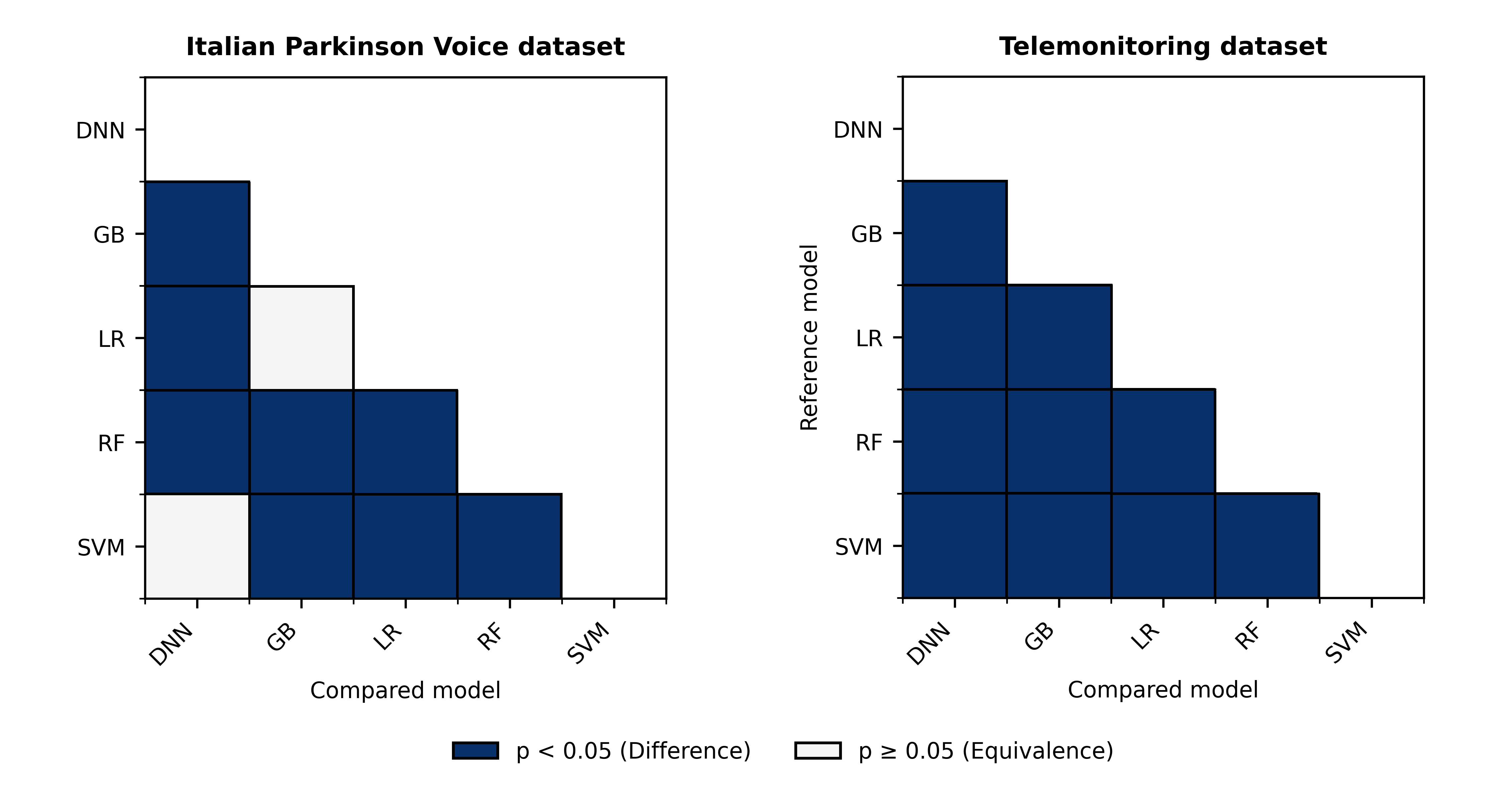}
\caption{Post-hoc pairwise comparison using Dunn test (Bonferroni correction).}
\label{fig:bonferroni}
\end{figure}

As previously discussed regarding the values in Table~\ref {tab:metrics_combined_datasets}, the DNN maintained the best accuracy, precision, and F1-score values in both datasets. However, in the Italian Voice data, all classifiers maintained similar performance, with an absolute performance difference no greater than $3.5$ percentage points. Therefore, to assess whether these disparities are statistically significant, in-depth inferential analyses were performed, as can be seen in Figure~\ref{fig:bonferroni}.

Since the assumptions of normality (Shapiro-Wilk) for the classifier accuracies and homoscedasticity (Levene) were not met, both with $(p<0.001)$, nonparametric methods were used. Furthermore, in Table~\ref{tab:global_tests}, it can be seen that the Kruskal-Wallis test confirmed that there are statistically significant differences between the models $(p<0.001)$ in both situations. 
\begin{table}[H]
\centering
\caption{Results of the normality, homogeneity, and overall comparison tests for both datasets.}
\resizebox{1\textwidth}{!}{
\begin{tabular}{lccc}
\hline
\textbf{Dataset} & \textbf{Normality (Shapiro–Wilk)} & \textbf{Homogeneity (Levene)} & \textbf{Kruskal–Wallis} \\
\hline
\textbf{Parkinson Telemonitoring} & Non-normal (p$<$0.001) & Non-homogeneous (p$<$0.001) & Significant difference (p$<$0.001) \\
\hline
\textbf{Italian Parkinson Voice} & Non-normal (p$<$0.001) & Non-homogeneous (p$<$0.001) & Significant difference (p$<$0.001) \\
\hline
\end{tabular}
}
\label{tab:global_tests}
\end{table}

The Bonferroni post-hoc analysis showed that, in the Italian dataset, the accuracy of the DNN and SVM is statistically equivalent , as shown in Figure~\ref{fig:bonferroni}. Similarly, the LR and GB models were not statistically different. However, these pairs, when compared with the other algorithms, presented statistically different performances.
In the telemonitoring dataset, all models differed statistically $(p<0.001)$.

Thus, the inferential analyses reinforce the robustness and validity of the descriptive conclusions about the metrics. Thus, the superiority of the DNN as the most reliable classifier for detecting PD using biomarkers is not merely a product of sampling variation.

\section{Discussion}\label{sec:Discutions}

The experimental results demonstrated that the Deep Neural Network (DNN) consistently outperforms the majority of traditional machine learning models on both evaluated datasets. In particular, the accuracies of the DNN and SVM models were identical ($p = 1$, Figure~\ref{fig:bonferroni}), within the margin of error on the Italian voice dataset. The SVM model achieved a recall of $0.9897$, slightly higher than that of the DNN ($0.9895$). This result is consistent with the findings of \cite{Pourhomayoun2021, Vasquez2019}, who compared the performance of SVM and DNN models in biomedical signal classification tasks. In a clinical context, the use of either method tends to reduce the probability of false-negative diagnoses, a decision-making process widely explored in the literature \cite{Abdar2021, Ahmad2022, Reyes2023}. All other classifiers also performed well, demonstrating the excellent discriminatory quality of vocal biomarkers in the Italian voice dataset.

On the Parkinson's disease telemonitoring dataset, the DNN model and the GB model showed statistically similar but not statistically identical average accuracies. This result aligns with the studies \cite{Grinsztajn2022, Gorishniy2023}. The first explains the reason for the approximation of the results between the GB and DNN models, and the second analyzes the performance of the DNN and argues that the superiority of the DNN is not absolute and depends on the data.
Furthermore, the DNN, having demonstrated high accuracy and low variance across multiple runs, is proposed as a simple, stable, and robust model, a statement corroborated by the study \cite{Ballester2022}, in the detection of diverse and complex biomarkers associated with PD.

Tree-based models, such as Random Forest and Gradient Boosting, also showed competitive performance, albeit with slightly greater variability. This finding is consistent with the work of \cite{Smith2018}, which used the same Parkinson's Telemonitoring of Voice dataset. Similarly, the work of \cite{Sivaranjini2019} found that tree-based models outperformed linear models.

Finally, Logistic Regression exhibited inferior performance in nearly all metrics across both datasets, highlighting its limitations in capturing the nonlinear relationships present in vocal biomarker data. The study by \cite{Lones2019}, consistent with the results of this study, also utilized Parkinson's Voice Telemonitoring data and found that all tested nonlinear classifiers outperformed Logistic Regression.

The application of rigorous non-parametric statistical tests, including Kruskal-Wallis and Post-hoc (Bonferroni), confirmed that the performance differences between the models were statistically significant, further validating the reliability of the results. Moreover, the results reinforce the diagnostic potential of voice-based analysis, suggesting that deep learning architectures are particularly suitable for biomedical applications where data complexity and inter-individual variability are prominent.

Now, the results of the approach proposed in this work will be compared with other notable studies that used sophisticated methodologies and technologies. It was observed, in this study, that the DNN with a mean accuracy of around $98.65\%$, on the Italian Voice dataset and with similar performance on the Parkinson Telemonitoring, presented itself as a stable model proposal for different contexts. The most recent study by \cite{Singh2025} used advanced AI models combined with hemodynamic parameters and a multimodal approach for the detection of PD and obtained an accuracy of $96\%\pm 0.0091$. In contrast, \cite{malekroodi2024leveraging} used the VGG16/19, ResNet50, and Swin transformer architectures, in the detection of healthy and diseased individuals, and the highest precision achieved was $96.67\%\pm 4.71$. This article also used the same Italian Voice dataset. According to \cite{PintoNeto2024}, the classical ML models (SVM and GB) and the hybrid Ensemble Stacking Model ESM are promising tools in the detection of PD. According to the reported values, in this work, their DNN obtained an accuracy of $0.7210$. Furthermore, an ESM with an accuracy of $0.8449$ was analyzed, using a single database that was unified from three distinct datasets. These discussions were summarized in Table~\ref{tab:comparaçao}.

\begin{table}[h!]
\centering
\scriptsize
\caption{Comparison of studies using vocal biomarkers for Parkinson’s disease detection.}
\label{tab:comparaçao}

\begin{threeparttable}
\rowcolors{2}{gray!10}{white}
\begin{tabular}{@{}p{2.9cm} p{3.6cm} c p{3.6cm}@{}}
\toprule
\textbf{Study} & \textbf{Model} & \textbf{Accuracy ± SD (\%)} & \textbf{Notes} \\
\midrule
\textbf{This work} & DNN (Proposed) & \textbf{98.65 ± 0.91} & Best performance on both datasets analyzed \\
Malekroodi et al. (2024) & CNN-based DL & 96.67 ± 4.71 & Italian voice \\
Singh et al. (2025) & Multimodal AI & 96 (n/r) & Requires sensors beyond voice \\
Pinto Neto et al. (2024) & Classical ML & 72.10 ± 6.65 & Mixed datasets  \\
\bottomrule
\end{tabular} 
\begin{tablenotes}
\footnotesize 
\item \textbf{Note:} $\text{n/r} = \text{not reported}$
\end{tablenotes}
\end{threeparttable}
\end{table}

Although the results obtained in this study and the comparison with existing literature validate the DNN's performance, future studies need to use voice datasets closer to the real clinical scenario. Such datasets should be heterogeneous, encompassing a variety of equipment and environments, as well as multiple languages.

\section{Conclusion} \label{sec:Conclusions}

The Deep Neural Network consistently outperformed traditional machine learning models on two distinct datasets, exhibiting low variability across multiple iterations and achieving improved performance on metrics such as accuracy, precision, and the F1-score. Its efficacy was verified not only descriptively but also through non-parametric tests, demonstrating significant differences compared to the majority of the models.

It was found that the DNN can efficiently identify the complex and non-linear characteristics present in vocal biomarkers associated with Parkinson's disease. Although other models may appear similar to the DNN, none maintained its performance across both scenarios. Statistical analyses confirmed the significance of the observed performance improvements, reinforcing that the DNN is robust and not limited to a single dataset.

Thus, by demonstrating superior performance, low computational cost during inference, and competitive results compared to much larger and more complex multimodal AI models or DNNs, the DNN architecture discussed here could become a method for screening in the detection of Parkinson's disease. Furthermore, the DNN classifier can be implemented to reach a large number of people in a non-invasive way, making it a strong candidate for the early detection of PD without the need for a specialized clinical environment.

\section*{Declarations}

\noindent\textbf{Ethics approval:} Not applicable.\\
\textbf{Funding:} This research was not funded externally.\\
\textbf{Competing interests:} The authors declare no conflicts of interest.\\
\textbf{Author contributions:} Katia Pires Nascimento do Sacramento : software, data analysis and writing. Elliot Q. C. Garcia: data analysis and writing. Nicéias Silva Vilela: data analysis and writing. Vinicius P. Sacramento: data analysis and writing. Tiago A. E. Ferreira: conceptualization, methodology and review.\\
\textbf{Data availability:} The UCI dataset is publicly available at: \url{https://archive.ics.uci.edu/dataset/186/parkinsons+telemonitoring}. The Italian dataset is located at: 
\url{https://ieee-dataport.org/datasets/italian-parkinsons-voice-and-speech}.

\section*{Declaration of generative AI and AI-assisted technologies in the manuscript preparation process}

During the preparation of this work, the authors used ChatGPT and Gemini in order to review the grammar rules of the English language in the manuscript. After using this tool/service, the authors reviewed and edited the content as needed and take full responsibility for the content of the published article.


\end{document}